\documentclass[a4paper]{article}
\usepackage{hyperref}

\usepackage{iwslt18,amssymb,amsmath,epsfig}
\usepackage{latexsym}
\usepackage{booktabs}
\usepackage{url}

\title{BPE and CharCNNs for Translation of Morphology: \\ A Cross-Lingual Comparison and Analysis}

 \makeatletter
 \def\name#1{\gdef\@name{#1\\}}
 \makeatother
 \name{{\em Pamela Shapiro, Kevin Duh}}

\address{Johns Hopkins University \\
{\small \tt pshapiro@jhu.edu,kevinduh@cs.jhu.edu}
}
\begin{document}
\maketitle
\begin{abstract}
Neural Machine Translation (NMT) in low-resource settings and of morphologically rich languages is made difficult in part by data sparsity of vocabulary words. Several methods have been used to help reduce this sparsity, notably Byte-Pair Encoding (BPE) and a character-based CNN layer (charCNN). However, the charCNN has largely been neglected, possibly because it has only been compared to BPE rather than combined with it. We argue for a reconsideration of the charCNN, based on cross-lingual improvements on low-resource data. 
We translate from 8 languages into English, using a multi-way parallel collection of TED transcripts. We find that in most cases, using both BPE and a charCNN performs best, while in Hebrew, using a charCNN over words is best.
\end{abstract}


%
\section{Introduction}
For efficiency reasons, standard neural machine translation models require a fixed vocabulary size, where rare words are treated as out-of-vocabulary. Even when it is possible to use the whole vocabulary, rare word embeddings are infrequently updated. To alleviate this sparsity problem, Byte-Pair Encoding~\cite{P16-1162} maps the vocabulary to subword units. This has been shown to improve translations of names, foreign words, and morphologically complex words, and has become standard practice~\cite{denkowski2017stronger}. 

Meanwhile, incorporating a character-based convolutional neural network (CNN) layer into the encoder~\cite{kim2016character, P16-2058} has also shown improvements in BLEU over standard sequence-to-sequence models. It alleviates sparsity on the source side by building word representations from characters. 


While other morphological practices have been suggested, BPE and charCNN are simple ways of managing morphological complexity. However, it is unclear to what extent their strengths are \textit{complementary}, how they might \textit{interact}, and where future efforts would best be directed.


We perform an empirical analysis across 8 languages, using a multi-way parallel test set from TED Talks.
Our results indicate that a charCNN is consistently part of the best solution for handling source morphology in this setting. It is usually best to use both BPE and a charCNN. However, BPE only marginally helps Arabic and harms Hebrew. This reinforces the argument for considering typology in natural language processing~\cite{bender2009linguistically}.

In addition to our analytical and empirical contributions, we provide an implementation of charCNN in OpenNMT-py.\footnote{\url{https://github.com/pamelashapiro/OpenNMT-py}}

\section{Methods Compared}

\subsection{Byte-Pair Encoding.}
Byte-Pair Encoding~\cite{P16-1162} uses an information theoretic algorithm from~\cite{gage1994new}. It initializes a symbol vocabulary to characters, then iteratively merges frequent symbol pairs, not allowing merges across word boundaries. The number of ``merge operations'' is the number of iterations in this algorithm. BPE has become standard practice and has been promoted as a technique for stronger baselines in NMT~\cite{denkowski2017stronger}. Note that while our analysis is focused on the source side, BPE also reduces the vocabulary on the target side.

\subsection{Character-Based CNN.}
\cite{kim2016character} introduced a language model that builds input word representations with a CNN architecture over characters. They run a convolutional layer over concatenated character embeddings, with max-pooling-over-time and a highway network, to form input word representations, which are then fed into an LSTM to make word-level predictions. \cite{P16-2058} incorporated this model into the NMT encoder and achieved improvements over standard seq2seq models. The charCNN layer is built for input to an RNN, so it is not as useful in the decoder~\cite{P17-1080}. Nevertheless, it can handle source-side morphology, which is the focus here.

\subsection{Conceptual Comparison.}
 BPE and charCNN approach sparsity reduction from different directions. By segmenting based on frequency, BPE keeps frequent words intact and is able to learn embeddings for subword pieces that might be useful. In addition, an attention mechanism can attend to any of these subword pieces. Meanwhile, the charCNN can learn which subword patterns are useful {\it jointly} with the model. Thus, we might expect these approaches to be complementary, and the experiments support this hypothesis.

\section{Experiments}

\begin{figure}
\includegraphics[width=0.50\textwidth]{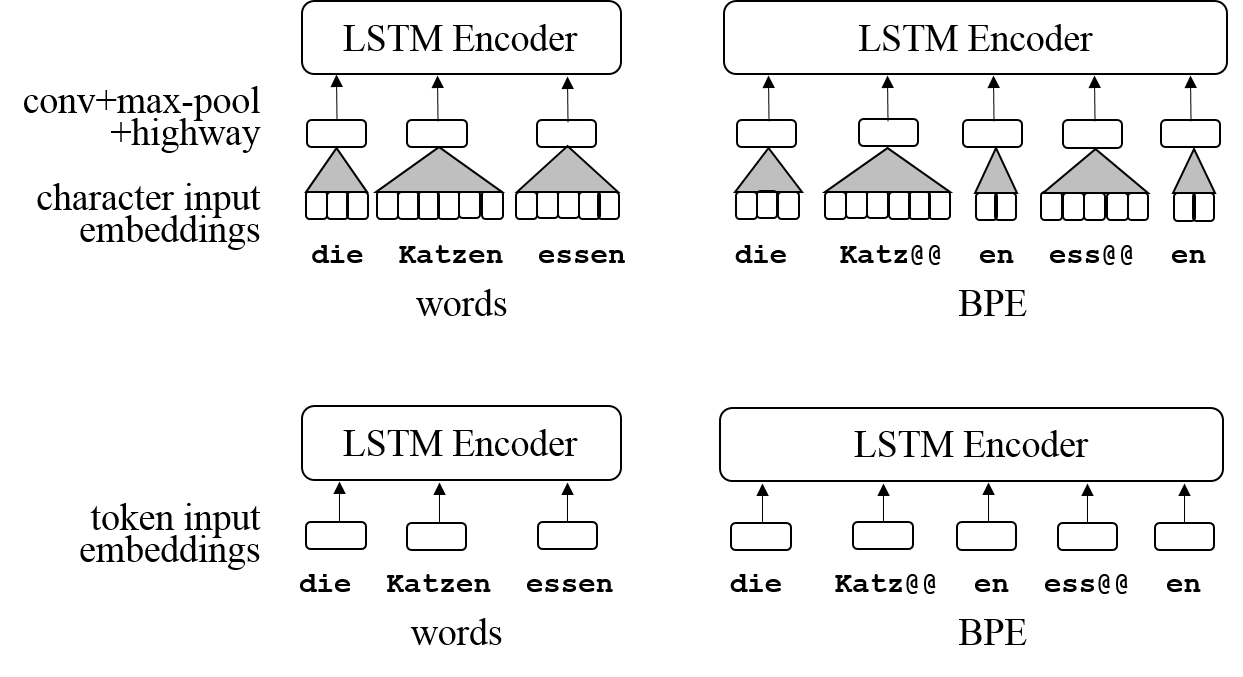}
\centering
\caption{\label{charCNN} Four of the different external processing (\texttt{word} or \texttt{bpe}) and model configuration (\texttt{tok} or \texttt{char}) combinations we examine in this work.}
\end{figure}

We illustrate the main cross-section of models and source processing we are comparing in Figure \ref{charCNN}.

\subsection{Dataset}

\begin{table}[ht]
\begin{center}
\begin{tabular}{lllll}
\toprule
\bf Language & \bf Tokens & \bf Types & \bf Sentences \\ 
\midrule
Arabic (ar) & 2,813k & 151k & 175k\\
French (fr) & 3,346k & 66k & 157k\\
German (de) & 2,850k & 107k & 152k\\
Hebrew (he) & 2,900k & 152k & 187k\\
Hungarian (hu) & 1,845k & 161k & 112k\\
Portuguese (pt) & 2,899k & 72k & 152k\\
Romanian (ro) & 2,999k & 97k & 161k\\
Russian (ru) & 3,120k & 149k & 180k\\
\bottomrule
\end{tabular}
\end{center}
\caption{\label{tab:data} Statistics of Training Data (8 source languages). Target is English (en). }
\end{table}

\begin{table*}
\begin{center}
\begin{tabular}{|c|c|c|c|c|c|c|}
\hline
& word+tok & word+char & bpe+tok & bpe+char & stem+tok & stem+char\\
\hline
de-en & 31.44    & 31.66     & 32.22   & \bf{32.74} & 29.87 & 30.86\\
\hline
fr-en & 34.79    & 35.39     & 35.02   & \bf{35.49} & 33.88 & 33.63\\
\hline
hu-en & 18.37    & 22.34     & 22.02   & \bf{22.62} & 17.99 & 18.95\\
\hline
pt-en & 39.79    & 41.14     & 40.92   & \bf{41.67} & 38.40 & 37.75\\
\hline
ro-en & 33.40     & 36.79     & 35.11   & \bf{36.97} & 33.69 & 34.62\\
\hline
ru-en & 22.41    & 23.74     & 23.33   & \bf{24.14} & 22.40 & 22.55\\
\hline
ar-en & 26.46    & 29.87     & 27.90   & \bf{29.93} & 22.02 & 22.08\\
\hline
he-en & 31.86    & \bf{35.53}     & 29.02   & 30.81 & N/A & N/A \\
\hline
\end{tabular}
\end{center}
\caption{\label{tab:bleu} BLEU scores comparing combinations of BPE and charCNN.} 
\end{table*}


\begin{table}
\begin{center}
\begin{tabular}{|c|c|c||c|c|}
\hline
& types & tokens & $\Delta_{word}$ & $\Delta_{stem}$\\
& stemmed & stemmed & & \\
\hline
ar & 11.50\% & 52.18\% & 3.41 & 0.06\\
\hline
fr & 33.88\% & 34.15\% & 0.60 & -0.25\\
\hline
de & 45.85\% & 32.22\% & 0.22 & 0.99\\
\hline
hu & 38.08\% & 44.14\% & 3.97 & 0.96\\
\hline
pt & 35.76\% & 43.75\% & 1.35 & -0.65\\
\hline
ro & 34.53\% & 36.96\% & 3.39 & 0.93\\
\hline
ru & 29.99\% & 51.93\% & 1.33 & 0.15\\
\hline
\end{tabular} 
\end{center}
\caption{\label{tab:stemmed} (left) Statistics for ISRI Stemmer for Arabic, and the Snowball Stemmer for the remaining languages. The percentage of types/tokens stemmed refers to which words in the training data were modified by the stemmer. (right) The improvement in BLEU charCNN makes over token-level seq2seq, showing charCNN is less helpful on stemmed text.}
\end{table}

As our dataset, we select 8 different source languages based on typological diversity from a multi-way parallel collection of TED talk transcripts.\footnote{\url{http://www.cs.jhu.edu/~kevinduh/a/multitarget-tedtalks/}} 
We setup 8 independent translation tasks with English as the target. 
Training data statistics for each task are in Table \ref{tab:data}. 
For all languages, the 2k sentence dev and test sets are multi-way parallel: this facilitates cross-linguistic comparisons on the resulting translations. 

\subsection{External Processing}

We experiment with 3 different types of sub-word processing. All versions begin with normalized and tokenized text.\footnote{For Arabic we normalize by removing diacritics and normalizing certain characters, as has been done in prior work.} Then, we experiment with no additional processing (\texttt{word}), Byte-Pair Encoding on the concatenation of source and target text (\texttt{bpe}), and for the sake of morphological analysis rather than a proposed alternative method, stemmed source text (\texttt{stem})\footnote{We used the Snowball Stemmer for 6 of the source languages. For Arabic, we use the ISRI Stemmer. For Hebrew, we could not find a stemmer.}. Our analysis uses 30k merge operations for BPE, though we also ran experiments with 100k merge operations and found performance was worse across the board. To compare morphological behavior, we do not truncate the source vocabulary.

\subsection{Model Hyper-Parameters}

We experiment with 2 different model variants, a standard seq2seq model at the token level (\texttt{tok}) and a model extended to have a charCNN layer (\texttt{char}). For both models, we use the default in OpenNMT-py of a bi-directional LSTM encoder-decoder model with the global, input-feeding, dot-product attention mechanism from ~\cite{D15-1166}. The hidden layer size is 1024, the batch size is 80, and the model was trained for 20 epochs with Adadelta~\cite{zeiler2012adadelta}, a dropout rate of 0.2, and a learning rate of 1.0. We use word-level prediction accuracy for model selection.

For the charCNN, we experimented with multiple kernel widths and found no improvement over using a kernel width of 6 for all 1000 convolutional kernels.\footnote{As is implemented in \url{https://github.com/harvardnlp/seq2seq-attn}} We used a maximum word length of 35 characters, and max-pooling and 2 highway layers as in~\cite{P16-2058}.

\section{Results}

\begin{table*}
\begin{center}
\begin{tabular}{|c|c|c|c|c|c|c|c|}
\hline
& NN & VB & IN & DT & PRP & JJ & RB \\
\hline
ar & 0.62\% & 2.06\% & 0.38\% & -0.64\% & 1.82\% & 2.36\% & 1.00\% \\
\hline
de & 0.62\% & 0.52\% & -0.38\% & 0.21\% & 0.84\% & 1.40\% & 0.80\% \\
\hline
fr & -0.68\% & 0.74\% & 1.16\% & -0.27\% & 0.72\% & 1.80\% & -0.60\% \\
\hline
he & 2.80\% & 	1.34\% &	-0.29\% &	-0.33\% &	1.03\%	& 3.12\% &	0.90\% \\
\hline
hu & 4.65\% & 3.29\% & 1.45\% & -0.67\% & 5.51\% & 2.32\% & 2.20\% \\
\hline
pt & 0.38\% & 1.31\% & 0.84\% & 0.09\% & 0.84\% & 0.28\% & 1.10\% \\
\hline
ro & 0.55\% & 0.97\% & 0.06\% & 1.00\% & 1.29\% & 1.52\% & -0.45\% \\
\hline
ru & 2.41\% & 1.52\% & 1.10\% & -0.85\% & -0.19\% & 2.16\% & -2.70\% \\
\hline
\hline
avg & 1.42\% & 1.47\% & 0.54\% & -0.18\% & 1.48\% & 1.87\% & 0.28\% \\
\hline
\end{tabular} 
\end{center}
\caption{\label{tab:pos} Percentage increase in recall of POS tags of English reference of \texttt{bpe+char} over \texttt{bpe+tok}.}
\end{table*}

In Table~\ref{tab:bleu}, we present the BLEU scores for all combinations of external processing and models. 
The general improvements of \texttt{bpe+char} over \texttt{word+char} and \texttt{bpe+tok} indicate that character modeling and BPE sub-units are complementary.
We also see that the charCNN is consistently part of the best solution for handling source morphology, while this is not always the case for BPE.
In particular, for Semitic languages, \texttt{bpe+char} is only marginally better than \texttt{word+char} in Arabic and substantially worse in Hebrew.\footnote{In experiments at 100k, BPE performed worse across the board, pushing \texttt{bpe+char} results to be worse than \texttt{word+char} for Hebrew, Arabic, and Romanian.}

By stemming source words, we can also observe to what extent our models are utilizing source inflectional morphology. Stemming generally hurts performance, indicating that our models are likely utilizing inflectional morphology to some extent. Moreover, we can look at to what extent the charCNN is able to improve translation in the word versus stemmed case. We show the improvements over token-level seq2seq in Table~\ref{tab:stemmed} with some statistics on the behavior of the stemmers for each language. In Hungarian, charCNN improves 3.97 BLEU over words and only 0.96 BLEU over stemmed text. This might imply that the charCNN is helping by improving representations of morphologically complex words, which it can no longer do in the stemmed case.

The strong performance of the charCNN is contrary to the charCNN's dismissal as not achieving competitive results.
This disconnect may be due to the absence of its implementation in the most up-to-date standard toolkits, or perhaps due to comparisons with highly-tuned BPE rather than combination with BPE. An additional misconception about the charCNN is that working with characters must be incredibly slow. However, our charCNN was only 10\% slower to train on GPU. It did consume more memory, but in particular for low-resource datasets such as this, it is not problematic.

\section{Qualitative Analysis}

It is difficult to know exactly what charCNNs are improving upon looking only at BLEU scores. To this end, we provide a more detailed analysis here.

\subsection{Translation of Various Parts of Speech}

To get a sense for what English words are translated better when a charCNN is added on top of BPE, over just BPE, we use a POS tagger on the reference and calculate recall for classes of POS tags. We provide the increase in recall in Table~\ref{tab:pos}. We see that personal pronouns, nouns, verbs, and adjectives see more improvement. These POS tags are often captured by inflectional morphology in other languages, indicating that the charCNN is capturing this morphology better than BPE alone. Meanwhile, recall of prepositions and determiners doesn't improve as much, and recall of determiners actually goes down slightly in many cases. Hungarian, a highly agglutinative language, shows the strongest differences in recall.

\subsection{Translation of OOV words}

We manually check translations of OOV words in the first 100 sentences of the test data for Arabic and Hebrew.
While this is a small sample size, the character models get OOVs more frequently. For example, in Arabic, the OOV 
{\it almtTwEp}, the word for ``the volunteer,'' gets translated correctly across all character systems and incorrectly across all token-level systems. It seems likely that this is because the verb for ``volunteer'' is a substring of this, 
{\it tTwE} and occurs in the training data.
In Hebrew, the OOV 
{\it lmiliSnivt}, the word for ``to milliseconds,'' gets translated correctly across character systems and incorrectly across token-level systems. Here, BPE breaks it up as {\it lmi@@ li@@ Snivt}, which separates the common word ``seconds'' out as the last word. However, this is not enough for the BPE token level system to get the word. It could be that the character-awareness here helps in particular with the transliteration of the English prefix ``milli-,'' which occurs consistently in other similar measure words in Hebrew.


\begin{table*}[ht]
\begin{center}
\begin{tabular}{|l|l|}
\hline
System & Output \\
\hline
\hline
src & dies ist ein bildungsprojekt , nicht ein \textbf{laptopprojekt} . \\
\hline
ref & this is an education project , not a \textbf{laptop project} . \\
\hline
word+tok & this is a crocodile , not a solo . \\
\hline
bpe+tok & this is a picture project , not a robust top project . \\
\hline
word+char & this is a project project , not a phantom project . \\
\hline
bpe+char & this is a picture project , not a \textbf{laptop project} . \\
\hline
\hline
src & welche davon k\"{o}nnten \textbf{selbstportraits} sein ? \\
\hline
ref & which ones of these could be \textbf{self-portraits} ? \\
\hline
word+tok & which of those could be in mind ? \\
\hline
bpe+tok & what might those of them could be portraits ? \\
\hline
word+char & which one of those could be suicide ? \\
\hline
bpe+char & which of those might be \textbf{self-portraits} ? \\
\hline
\end{tabular} 
\end{center}
\caption{\label{tab:outputs} Example German translation outputs for which bpe+char performs best.}
\end{table*}

\subsection{Semitic BPE Segmentations}
\label{language-specific}

BPE is least helpful for Arabic and actually harmful for Hebrew. Manually inspecting the source-side Hebrew test data, we see that some breaks are between morphemes, as in {\it l@@ rvfah}, ``to a doctor," where {\it l} is a prefix meaning "to."
At other times, breaks occur at bad locations, such as in the middle of {\it mk@@ ndh}, ``from Canada,'' causing it to no longer be able to produce the word Canada (where the \texttt{word+tok} system does). 
We observed a similar bad break in Arabic, 
{\it alkn@@ dywn}, ``the Canadians,'' where it breaks in the middle and can no longer translate it. One explanation for why these errors are more common in these languages is that they have ``impure abjad'' alphabets, meaning they do not standardly encode short vowels. This ties in with their templatic morphology, which builds words off of a core consonantal root. In these examples where the sequence of characters for ``Canada'' gets broken up, there is less identifying information left about the word than there would be in other languages (just the consonants).

\subsection{When does BPE+CharCNN do best?}

To understand further why the combination of BPE and CharCNN proves useful, we look at sentences where bpe+char outperformed the other systems in the German-English translations. We find an interesting phenomenon, which we provide examples of in Table \ref{tab:outputs}. In these instances, there is a compound word, which gets broken up by BPE. With bpe+tok, the second word in the compound gets correctly translated, while the first (which has @@ symbols attached as part of BPE), does not. CharCNNs seem to help resolve this issue. As for word+char, in the second example, we see evidence of modeling some of the word's spelling---the first part of the compound, {\it selbst-}, is also in the German word for ``suicide,'' {\it selbstmord}, which it is falsely translated as. It seems that in cases like this with long compound words, neither BPE nor charCNN alone is enough to model the morphology.

However, it is also possible that some of the gains associated with the character model are less interpretable. It is unclear how it affects optimization of the system as a whole, and this would be good to look into in future work.

\section{Related Work}

A few papers have worked with the charCNN and analyzed its performance.~\cite{P17-1080} show that the hidden states of the charCNN can classify morphology better than standard seq2seq models at the word level.~\cite{belinkov2017synthetic} show that charCNNs are sensitive to various types of noise. Our analysis complements this line of work, supporting the conclusion that the charCNN is learning morphology better than standard seq2seq models and explores additionally how its strengths interact with BPE.



\section{Conclusion}

There is compelling evidence that the charCNN improves translation of morphology in ways that are complementary to BPE. The architecture is not slow like an RNN over characters would be, and there is no reason the charCNN cannot be combined with other encoder-decoder architectures. Thus, we recommend the use of charCNN for stronger source-side morphological baselines in low-resource settings.
Additionally, it seems there is some morphology that BPE helps to capture in non-Semitic languages. Thus, future work should investigate alternatives to BPE for Semitic languages. Finally, all of this work focuses on the encoder, where the charCNN has seen gains. Additional future work should look into similar approaches for the decoder.

\bibliography{emnlp2018}
\bibliographystyle{IEEEtran}
\end{document}